\documentclass{article}

\usepackage{arxiv}

\usepackage[utf8]{inputenc} 
\usepackage[T1]{fontenc}    
\usepackage{hyperref}       
\usepackage{url}            
\usepackage{booktabs}       
\usepackage{amsfonts}       
\usepackage{nicefrac}       
\usepackage{microtype}      
\usepackage{lipsum}		
\usepackage{graphicx}
\usepackage{natbib}
\usepackage{doi}
\usepackage{tabularx}
\usepackage{multicol}
\usepackage{xcolor}
\usepackage{adjustbox}
\usepackage{booktabs}
\usepackage{multirow}
\usepackage{stfloats}

\makeatletter
\def\blfootnote{\xdef\@thefnmark{}\@footnotetext}
\makeatother

\title{Towards Social Situation Awareness in Support Agents}

\author{ \href{https://orcid.org/0000-0002-1662-6652}{\includegraphics[scale=0.06]{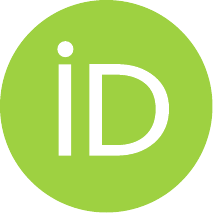}\hspace{1mm}Ilir Kola} \\
	Delft University of Technology\\
	The Netherlands \\
	\texttt{i.kola@tudelft.nl} \\
	\And
	\href{https://orcid.org/0000-0002-1261-6908}{\includegraphics[scale=0.06]{orcid.pdf}\hspace{1mm}Pradeep K.~Murukannaiah} \\
	Delft University of Technology\\
	The Netherlands \\
	\texttt{p.k.murukannaiah@tudelft.nl} \\
	\And
	\href{https://orcid.org/0000-0003-4780-7461}{\includegraphics[scale=0.06]{orcid.pdf}\hspace{1mm}Catholijn M.~Jonker} \\
	Delft University of Technology \& \\ Leiden Institute for Advanced Computer Science \\
	The Netherlands \\
	\texttt{c.m.jonker@tudelft.nl} \\

	\And
	\href{https://orcid.org/0000-0001-9089-5271}{\includegraphics[scale=0.06]{orcid.pdf}\hspace{1mm}M. Birna van Riemsdijk} \\
	University of Twente\\
	The Netherlands \\
	\texttt{m.b.vanriemsdijk@utwente.nl} \\
}

\date{}

\renewcommand{\headeright}{}
\renewcommand{\undertitle}{}
\renewcommand{\shorttitle}{Towards Social Situation Awareness in Support Agents}

\hypersetup{
pdftitle={A template for the arxiv style},
pdfsubject={q-bio.NC, q-bio.QM},
pdfauthor={David S.~Hippocampus, Elias D.~Striatum},
pdfkeywords={First keyword, Second keyword, More},
}

\begin{document}
\maketitle

\begin{abstract}
Artificial agents that support people in their daily activities (e.g., virtual coaches and personal assistants) are increasingly prevalent. Since many daily activities are social in nature, support agents should understand a user's social situation to offer comprehensive support. However, there are no systematic approaches for developing support agents that are social situation aware. We identify key requirements for a support agent to be social situation aware and propose steps to realize those requirements. These steps are presented through a conceptual architecture centered on two key ideas: (1) conceptualizing social situation awareness as an instantiation of `general' situation awareness, and (2) using situation taxonomies for such instantiation. This enables support agents to represent a user's social situation, comprehend its meaning, and assess its impact on the user's behavior. 
We discuss empirical results supporting the effectiveness of the proposed approach and illustrate how the architecture can be used in support agents through two use cases.
\end{abstract}

\section{Introduction}\label{sec:introduction}

Human behavior is a function of a person's characteristics as well as the situation \citep{lewin1939field}. Thus, to support people in their daily lives, artificial agents must represent and reason about not only the personal characteristics but also the situation of a user. To take a user's situation into account, support agents should reason about the user's surrounding entities and how they relate to each other and the user. \blfootnote{This article appears for publication at IEEE Intelligent Systems (doi: \href{http://dx.doi.org/10.1109/MIS.2022.3163625}{10.1109/MIS.2022.3163625}). © 2022 IEEE. Personal use of this material is permitted.  Permission from IEEE must be obtained for all other uses, in any current or future media, including reprinting/republishing this material for advertising or promotional purposes, creating new collective works, for resale or redistribution to servers or lists, or reuse of any copyrighted component of this work in other works.}

Most of our daily situations are social in nature. We collaborate with co-workers, spend weekends with family and friends, and share most of our moments with people. Thus, support agents should account for this social dimension of our lives. 

We define social situation awareness and propose the building blocks necessary for support agents to be social situation aware. These building blocks are presented through a conceptual architecture inspired by work on `general' situation awareness \citep{endsley1995toward}, which we instantiate with concepts from social sciences \citep{rauthmann2014situational} to account for the requirements of modelling social situations. This work serves as a proof of concept showing that the building blocks of social situation awareness can be implemented in support agents and discusses the remaining steps for successful deployment of a full-fledged agent. 

\section{WHAT IS SOCIAL SITUATION AWARENESS?}

\cite{yang2009concept} define a situation as ``a  combination of the individually interpreted, implicit, and unique understandings, and the culturally shared, explicit, and common understandings of the surroundings that produce and constrain human behavior." We define a \emph{social situation} as a type of situation that involves more than one person. Thus, a social situation involves not only the typical situational elements such as time and place but also social elements such as the quality of the relationships and contact frequency between the user and other people in the situation. 

The social elements of a situation influence user behavior. For instance, consider two situations: one in which a user has dinner with a friend and another in which the user has dinner with a prospective employer. In these two situations, despite similar environmental elements, the user's behavior can be different because of the different relationships among the people in these situations. 

\cite{endsley1995toward} describes a prominent model of situation awareness consisting of three levels: (1) \emph{perception}, representing the status, attributes and dynamics of relevant elements in the environment; (2) \emph{comprehension}, representing a higher level synthetized meaning of the elements of the environment; and (3) \emph{projection}, representing the ability to project the future status of the elements of the environment. Adapting Endsley's model, we define \emph{social situation awareness} as: 

\begin{quote} A support agent's ability to perceive the social elements of a situation, to comprehend their meaning, and to infer their effect on the behavior of the user.\end{quote} 

\section{SITUATION TAXONOMIES}

Situations are abstract entities, which makes assigning meaning to them challenging. Studies in social psychology \citep{rauthmann2014situational} suggest that people interpret situations by creating impressions of them as if they were real entities which have specific \emph{psychological characteristics}. Understanding situations in terms of these characteristics allows people to better navigate the world by using these characteristics to predict what will happen and coordinate behavior accordingly. We propose that support agents should similarly treat situations as real entities with psychological characteristics.

Psychological characteristics provide a high-level subjective interpretation of situations and are widely studied, and different taxonomies have been developed. Here we present the elements of the DIAMONDS taxonomy \citep{rauthmann2014situational}. We choose this taxonomy because it is designed to cover daily situations and it offers a validated scale for measuring the psychological characteristics of situations. The taxonomy comprises the following characteristics in terms of which situations can be described:

\begin{itemize}
    \item \textbf{Duty} - situations where a job has to be done, minor details are important, and rational thinking is called for;
    \item \textbf{Intellect} - situations that afford an opportunity to demonstrate intellectual capacity;
    \item \textbf{Adversity} - situations where you or someone else are (potentially) being criticized, blamed, or under threat;
    \item \textbf{Mating} - situations where potential romantic partners are present, and physical attractiveness is relevant;
    \item \textbf{pOsitivity} - playful and enjoyable situations, which are simple and clear-cut;
    \item \textbf{Negativity} - stressful, frustrating, and anxiety-inducing situations;
    \item \textbf{Deception} - situations where someone might be deceitful. These situations may cause feelings of hostility;
    \item \textbf{Sociality} - situations where social interaction is possible, and close personal relationships are present or have the potential to develop.
\end{itemize}

\cite{rauthmann2014situational} suggest that people use these psychological characteristics to ascribe meaning to situations. Furthermore, they show that psychological characteristics of situations correlate with various situation cues, as well as behaviors that people exhibit in those situations. For instance, a high level of Duty is characteristic of work situations, and typical behaviors for situations with a high level of duty are concentrating and displaying ambition. This corresponds to our definition of social situation awareness: psychological characteristics of situations can be used for social situation comprehension, and are related to both social situation perception and social situation comprehension.

\section{REQUIREMENTS FOR SOCIAL SITUATION AWARE AGENTS}

Different context awareness architectures have been proposed for different purposes. \cite{alegre2016engineering} provide a review of existing approaches, and suggest that one of the reasons for the variety of existing approaches is the need for specific architectures in each domain. However, none of the reviewed approaches tackles social situations specifically. Our research fills this gap. Focusing on social situations motivates us to take into account the human aspects of a situation as opposed to the technical aspects investigated in related work, such as geo-spatial locations and other physical elements of context. Furthermore, the focus of existing approaches is on information that can be acquired through sensors, which is processed to determine actions that are occurring in the environment. Our work complements these approaches by focusing on the psychological meaning of situations. Based on these differences, we formulate the following requirements for support agents to be social situation aware.

\bigskip
\noindent \textbf{Combining sensory data with a user's mental constructs}
Perceiving social situations relies not only on information that can be detected through sensors, but also on a user's mental constructs. For instance, in a situation where a user is meeting another person for dinner it is difficult to detect the features of their relationship from sensors alone. This information can be important, e.g., a dinner with a friend is very different from a dinner with a potential employer. Therefore, the agent needs to be able to elicit information about the user's mental constructs such as social relations, which may not be available via sensors.

\bigskip
\noindent \textbf{Variety of social situations}
A flexible support agent should be able to represent a wide variety of social situations a user may encounter. To do so, an agent must identify a variety of social dimensions characterizing a situation. Further, the agent should be able to interpret this situation variety by translating social features into abstractions to determine appropriate support, e.g., using pre-specified rules to categorize situations into a limited number of higher-level situations, or using machine learning to derive information that can be used for reasoning about support.

\bigskip
\noindent \textbf{Interpreting the meaning of situations}
Existing work on situation awareness addresses the comprehension step by determining how the perceived objects in a situation are interrelated \citep{baumgartner2006survey} and recognizing the situation type. For instance, if two users are perceived in the same office, the comprehension step would say that the user is in a meeting. However, in social situations, knowing the type of situation is not sufficient to determine the support needed since it is possible to infer different meanings from this information. For example, being in a meeting with a supervisor is different from a meeting with a potential client. Support agents need to be able to distinguish the different meanings of such social situations.

\bigskip
\noindent \textbf{Value-aware support}
Agents should provide support that is consistent with the user's goals and preferences. In social science, it has been argued that the essence of a situation is its affordance of human goals and motives \citep{rauthmann2014situational}. A way to represent human motives are personal values. Values such as independence or success which express what people find important in life have been found to be key drivers of human decisions, and value preferences exhibit cross-situational consistency \citep{schwartz1992universals}. Since providing support in social situations is ultimately about aligning with the user's underlying motivations, we suggest the use of values for personalization.

\bigskip
\noindent \textbf{Explainability and directability} 
Support agents need to be able to explain their suggestions to users. For instance, consider an agent that supports healthy lifestyle. If the agent merely suggests the user to avoid going to a party, this advice might be unexpected. However, if the agent informs the user that going to parties usually leads to smoking, which demotes the value of `health', then the user can make an informed decision. Further, the user should be able to direct the agent on how to act. Continuing our example, the user should be able to inform the agent that the party is in a non-smoking venue, health would not be demoted. The agent can then use this information in future situations.

\bigskip
Although variants of these requirements are mentioned in existing work, to the best of our knowledge our formulation and approach towards tackling them in an integrated way is new. The key novel elements in our requirements are the consideration of how to ascribe meaning to social situations, the emphasis on user interaction, and a hybrid human-machine approach for social situation awareness and support.

\section{A CONCEPTUAL ARCHITECTURE FOR SOCIAL SITUATION AWARENESS}

We identify the core elements, and their interrelations, for creating social situation aware agents by presenting a conceptual architecture. The architecture consists of two main components: a social situation awareness component, and a user interaction component. The first is an instantiation of the three-level situation awareness model proposed by \cite{endsley1995toward} with social concepts. The second comprises interaction modules needed for integrating situation awareness reasoning with the supportive function of the agent. We provide directions for implementing these components.

\begin{figure}[!htb]
    \centering
    \includegraphics[width=\textwidth]{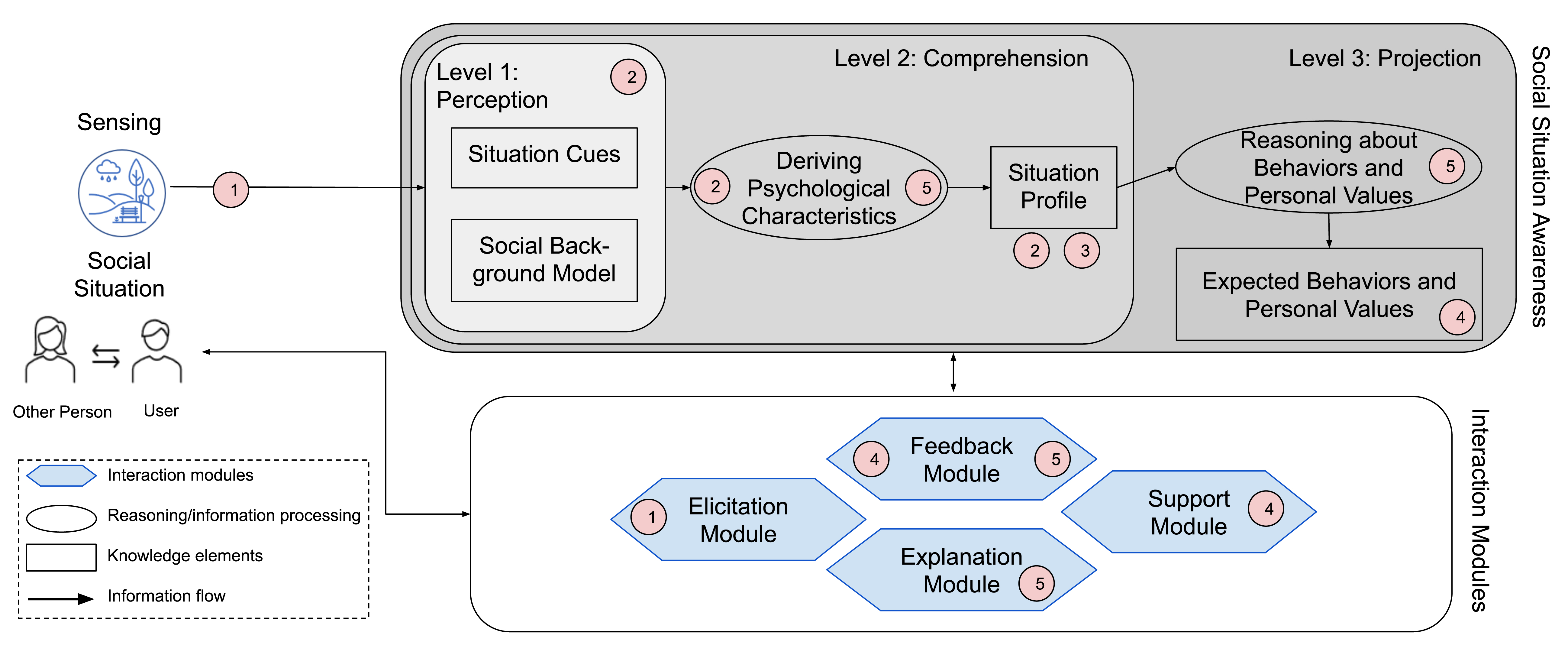}
    \caption{Architecture of a social situation aware support agent. The numbers in red circles represent the requirements (Table~\ref{tab:requirements-mapping}) that the elements of the architecture address.}
    \label{fig:ssa}
\end{figure}

\subsection{Level 1: Social Situation Perception}

The goal of the perception level is to obtain a representation of the salient aspects of a social situation. This information can come from sensory data and interaction with the user. To account for a wide variety of situations, the information included in this level should allow representing arbitrary social situations. \cite{kola2019s} propose an approach to model arbitrary social situations through a two-level ontology distinguishing \emph{situation cues} and social relationship features (\emph{social background model}). \cite{rosatelli2019detecting} propose an approach where data from wearable sensors is processed with deep learning techniques to assess information such as roles in social interactions.

\subsection{Level 2: Social Situation Comprehension}
In this level, the perceived elements are used to infer a social situation profile, characterizing the situation along meaningful dimensions.

\noindent\textbf{Knowledge elements}
To describe the meaning of a social situation, we propose to use the psychological characteristics of situations (see section on situation taxonomies). The idea is to describe each social situation by a set of features (the \emph{situation profile}) that represent the psychological characteristics of that situation. These characteristics describe a user's subjective understanding of a situation. A key advantage of this approach is that it offers a fixed number of dimensions based on which it is possible to represent and compare different situations. 

\noindent\textbf{Reasoning}
To determine the psychological characteristics of a situation, one may follow a rule-based or a machine learning approach. A rule-based approach provides explicit reasoning, but requires extensive design time specifications. A machine learning approach supports situation variety, e.g., by offering predictions for unseen examples, but offers limited explainability. 

\subsection{Level 3: Social Situation Projection}
In this level, the agent uses the situation profile to predict how the user is likely to behave in a social situation, and what values are affected.

\noindent\textbf{Knowledge elements}
In the classic situation awareness model, the projection level captures how the situation develops as a whole. To fulfill the personalization requirement, we propose that in the projection level the agent needs to predict what behavior the user is likely to exhibit, and the personal values promoted or demoted in a given situation. The former allows the agent to provide proactive support, and the latter enables the agent to help the users in a value-aligned manner. 

\noindent\textbf{Reasoning}
This component takes the situation profiles as input, and predicts the expected behavior and the promoted and demoted values. A possible way to do this is by grouping similar situations based on their profile, and studying the patterns of behaviors and values in each group of situations, as done by \cite{kola2020grouping}.

\subsection{Interaction Modules}
An agent needs to interact with the user to give and acquire information necessary for support. We foresee the need for four interaction modules. In this paper, we focus on describing the role of these modules as part of the envisaged support agent. 
In order to realize the interaction modules and create a full-fledged social situation aware agent, open research challenges regarding human-machine hybrid intelligence \citep{akata2020research} need to be addressed. 

\subsubsection{Elicitation Module}
The elicitation module interacts with the user to elicit necessary information that cannot be acquired from a sensor. User interaction is needed during both initialization and run time. During initialization, the goal is to gather information that remains relatively stable, e.g., information about a user's social relationships with their most frequent contacts, needed to form the social background model. This ensures that for most social situations which the user encounters, the social background model already contains the needed information, thus avoiding to overload the user with information requests after initialisation. During run time, the module detects when certain information is missing regarding a specific social situation, e.g., the role of the other person, and asks the user. The Platys framework \citep{murukannaiah2015platys} can be used to reduce the possible burden of information elicitation for the user. Platys employs an active learning approach, which asks a user to provide context information only if the predictions with existing sensor readings are uncertain, which reduces the overall effort for the user.

\subsubsection{Support Module}
After going through the social situation awareness levels, an agent can reason about the support it can provide. One of the proposed requirements is for the agent to personalize support according to the needs and the values of the user. This information can be contained in a user model within the support module. The support module can then compare the user preferences with the information coming from Level 3 of the architecture regarding expected user behavior and values. Support is needed when there is a mismatch between the preferred and the expected behavior of the user in a situation, or when the situation affects a value important to the user. 

\subsubsection{Explanation Module}
To make an agent's support actions explainable, we propose to use meaningful social notions in each level of the architecture, derived through explainable reasoning and learning techniques. An advantage of a multi-level architecture is that explanations can be given on different levels: the agent can (1) give insight on the suggestion relating it to a certain personal value or preferred behavior (Level 3), (2) explain why a certain behavior or personal value is expected in a specific social situation by referring to the psychological characteristics of that situation (Level 2), and (3) give further insight on the situation cues and social relationship aspects that cause the situation to have those specific psychological characteristics (Level 1). 

\subsubsection{Feedback module}
It should be possible for the user to notify the agent when a support action or its explanation is not satisfactory. The feedback module achieves this by interacting with the user to determine whether there has been a mistake in one of the reasoning steps or whether some information in the knowledge base needs to be updated. The agent can then integrate this feedback into its reasoning mechanisms and knowledge bases at the appropriate level. How exactly such updates are to be performed and represented is an open research question. 

\section{EMPIRICAL EVIDENCE}

In this section, we present empirical evidence that supports our proposed three-level social situation awareness component. The social situation awareness component is an instantiation of the well-known model of situation awareness by \cite{endsley1995toward}. The model's diverse applications suggest that the three level approach as a whole is beneficial. 

\begin{table}[!h]
\caption{Key requirements and how they are addressed in our proposed approach}
\resizebox{\textwidth}{!}{\begin{tabular}{lll}
\hline
\textbf{Requirement}                                               & \textbf{How it is addressed}                                                                                                                                                                          & \textbf{Empirical evidence}                                                   \\ \hline
1) Combining sensory data \\ with mental constructs of the user & Perception level based on sensory data and user-elicited information                                                                                                                         & \cite{kola2019s}                                    \\ \hline
2) Variety of situations                                     & \begin{tabular}[c]{@{}l@{}}Use concepts from social sciences to allow representing arbitrary situations\\ Use machine learning to learn connections between Level 1 and Level 2\end{tabular} & \cite{rauthmann2014situational, kola2019s}          \\ \hline
3) Interpreting the meaning of situations                    & Derive the psychological characteristics of situations                                                                                                                                       & \cite{rauthmann2014situational, kola2021predicting} \\ \hline
4) Value-aware support                                       & \begin{tabular}[c]{@{}l@{}}Base support on the personal values of the user\\ Have feedback module which allows personalization\end{tabular}                                                  & \cite{kola2020grouping}                             \\ \hline
5) Explainability and directability                          & \begin{tabular}[c]{@{}l@{}}Use explainable techniques\\ Explanation module\\ Feedback module\end{tabular}                                                                                    & \cite{kola2021predicting}                           \\ \hline
\end{tabular}}
\label{tab:requirements-mapping}
\end{table}

In past work, the different levels of the social situation awareness were implemented and evaluated through human-grounded studies \citep{doshi2017towards}. Human-grounded evaluations involve real people who are presented with simplified tasks, and are particularly useful in cases such as ours where the goal is to evaluate reasoning components and a full-fledged agent cannot yet be implemented due to open challenges in interaction modules. In \cite{kola2021predicting}, we showed that transitioning through the three levels of the architecture is possible: using data collected from a large user study, we presented an approach in which it is possible to predict Level 2 information from Level 1 inputs, and then in turn use the predicted Level 2 information as input for predicting Level 3 information. Furthermore, we showed how Level 1 and Level 2 information can be used as a basis for creating explanations that are satisfying for people. In this section we give details on how the different levels of the social situation awareness module have been successfully implemented and evaluated in generic domains, e.g., to assess the promoted or demoted personal values of a social situation, or specific domains, such as reasoning about the priority of social situations. Furthermore, in Table~\ref{tab:requirements-mapping} we present how each architectural element tackles the identified requirements.   

\noindent\textbf{Level 1} In \cite{kola2019s}, we proposed an ontology to tackle the perception level. The ontology models situation cues, describing the situation, and social relationship features, describing the relationship of the user with the people in the situation. We evaluated this approach via a user study in which participants were asked about their social relationships using the features proposed in the ontology. Participants considered the ontology to contain an appropriate amount of information (average answer=3, SD=0.61 on a 5 points Likert scale where 1=too little information, 3=appropriate information, 5=too much information) and to be fairly representative of their social relationships (average answer=3, SD=0.79 on a 5 points Likert scale where 1=very little representative and 5=very much representative). This study suggests that it is possible to have a model of a social situation that includes a user's mental constructs, in particular describing social aspects of situations, thus fulfilling Requirement 1.

\noindent\textbf{Level 2} In this level, we suggest ascribing meaning to situations through the psychological characteristics. \cite{rauthmann2014situational} conducted validation studies involving hundreds of participants across different countries and cultures, showing that the DIAMONDS taxonomy can be used to give meaning to arbitrary situations, thus providing evidence for Requirements 2 and 3. A technical requirement of the architecture is the ability to derive these psychological characteristics from the information from Level 1. To investigate this, we collected Level 1 and 2 data through a crowdsourcing user study \citep{kola2021predicting}. Using this data, we showed that machine learning models can be created that predict psychological characteristics of situations from Level 1 information with an average error of 1.14 on a 6-point Likert scale, outperforming benchmark results.

\noindent\textbf{Level 3} In \cite{kola2020grouping}, we proposed an approach that groups situations based on psychological characteristics and show that different personal values are promoted or demoted in specific groups of situations. For instance, we noticed that situations with high intellect and duty promote the values helpfulness and capability. This helps fulfilling Requirement 4. Further, this shows that transition from Level 2 to 3 is feasible with respect to personal values. To show that this transition is also possible in terms of expected behaviors, in \cite{kola2021predicting} we used psychological characteristics of situations as input to predict expected user behavior regarding social priorities  with an error of 0.98 on a 7-point Likert scale for actual values of the characteristics, and with an error of 1.37 for predicted values of psychological characteristics based on Level 1 information.

\section{USE CASES}

We illustrate how the components of our approach could be included in intelligent agents that provide support via two use cases: agenda management \citep{kola2021predicting} and location data sharing \citep{kayal2018automatic} support agents. Although these use cases are quite different, the high-level components of our approach can be instantiated for each use case as shown in Table~\ref{tab:use_cases}. This illustrates how our approach can serve as a blueprint for including social situation awareness in support agents.

\begin{table}[!htb]
\centering
\caption{Concepts that can be modelled and role of modules in two use cases.}
\begin{tabular}{@{}l>{\raggedright}p{6.8cm}>{\raggedright\arraybackslash}p{6.8cm}@{}}
\toprule
\textbf{Use Case} & Agenda Management Support Agent & Value-based Location Sharing Agent \\
\midrule\midrule
\textbf{Level 1} & Social background features of other person (e.g., role, hierarchy level) & Location-related features; Other people present \\
\midrule
\textbf{Level 2} & Psychological characteristics of situation (e.g., Duty, Intellect) & Psychological characteristics of situation (e.g., Sociality) \\
\midrule
\textbf{Level 3} & Predict priority of meetings & Assess how values are affected \\
\midrule
\textbf{Elicitation} & Social situation features & Personal values \\
\midrule
\textbf{Support} & Suggest which meeting to attend based on priority & Provide value-aligned support \\
\midrule
\textbf{Explanation} & Why a meeting was suggested & Why a location was shared with someone \\
\midrule
\textbf{Feedback} & Adapt priority prediction model & Adapt value assessment \\
\bottomrule
\end{tabular}
\label{tab:use_cases}
\end{table}

\subsection{Agenda Management Support Agent}
\cite{kola2020predicting} introduce an agenda management support agent, whose goal is to assess a user's priorities and make suggestions based on the priority levels when different meetings overlap.
Table~\ref{tab:use_cases} illustrates the information modelled in the different components of the architecture. Level 1 (\emph{perception}) includes information such as the role of the other person and their hierarchy level. The agent uses the perceived information to assess the psychological characteristics of the situation, which are modeled through the DIAMONDS taxonomy (\emph{comprehension}). From this information, the agent determines a priority level for every social situation (\emph{projection}). If two meetings overlap, the agent suggests to the user to attend the meeting with higher priority and reschedule the other. The user can ask for the reason behind the suggestion, and explanations can be given based on Level 2 or Level 1 information. If the user does not accept the suggestion, the \emph{feedback} module asks about the reasons and incorporates the feedback into the knowledge base and reasoning processes to better predict priority in similar future situations.

In this use case, following our approach  allows to explicitly take into account social aspects of the situation, which are modelled from the point of the view of the user through the elicitation module and the perception level. Furthermore, our proposed situation comprehension approach allows for a richer representation and understanding of situations, which in turn allows to better assess priority. For instance an agent may determine that meetings involving a high level of duty are more important for a specific user.

\subsection{Value-based Location Sharing Agent}
\cite{kayal2018automatic} propose a model for choosing among conflicting agreements about social sharing of location data based on the users' personal values. They show that an agent can help in resolving conflicting commitments by knowing the value preferences of the user and the promoted values of different location sharing commitments. 

Our social situation awareness framework can extend this approach. Level 3 (\emph{projection}) enables the agent to automatically assess which values are promoted or demoted in a situation. Once this information is available, the \emph{support} module can assess whether a specific location sharing activity is aligned with the values of the user. Including information about social relations (Level 1) allows a prediction of values based on a richer model. Furthermore, explicitly modeling the psychological characteristics of the situation (\emph{comprehension}) can be beneficial since these have been shown to be a good predictor of personal values afforded in a situation \citep{rauthmann2014situational, kola2020grouping}. For instance, the agent may infer that situations taking place in specific locations involve a high level of sociality, and such situations also tend to promote the value social recognition. If the value is important for the user, the agent would share the location data. This information would also facilitate \emph{explanations}: if the user asks why the location was shared, the agent would explain that it had inferred that the situation promotes social recognition because it involves a high level of sociality. If this inference is not correct, the \emph{feedback} module would adapt the value assessment model accordingly.

\section{CONCLUSION}

We outline the elements needed for social situation awareness in support agents and illustrate their practical benefits. Existing work (e.g., \citep{kola2020grouping, kola2020predicting, kola2021predicting}) has shown promising results in implementing the different levels of social situation awareness, as well as in automatically transitioning between the levels using data from studies conducted with real people. This work serves as a proof of concept for social situation awareness in support agents. However, more research from different communities is needed to go from this proof of concept to a full-fledged agent. Firstly, the interactive modules will have a crucial role in the successful implementation of an agent that can be tested on real tasks with users. Realizing these requires further research investigating how hybrid intelligent systems can be made collaborative, adaptive, responsible, and explainable \citep{akata2020research}. This includes advances  in integrating active learning approaches in order to better personalize the prediction models for specific users based on their feedback. Lastly, this proposed architecture should be integrated with work on interpreting the meaning of social signals such as body language in social interactions \citep{rosatelli2019detecting}. This would allow the agent to take into account the dynamics of a social situation as it unfolds, allowing it to integrate social situation understanding based on social relations with observed social signals.

\section*{Acknowledgement}

This work is part of the research programme CoreSAEP, with project number 639.022.416, which is financed by the Netherlands Organisation for Scientific Research (NWO).

\bibliographystyle{plainnat}
\bibliography{references}  

\begin{thebibliography}{16}
\providecommand{\natexlab}[1]{#1}
\providecommand{\url}[1]{\texttt{#1}}
\expandafter\ifx\csname urlstyle\endcsname\relax
  \providecommand{\doi}[1]{doi: #1}\else
  \providecommand{\doi}{doi: \begingroup \urlstyle{rm}\Url}\fi

\bibitem[Akata et~al.(2020)Akata, Balliet, de~Rijke, Dignum, Dignum, Eiben,
  Fokkens, Grossi, Hindriks, Hoos, et~al.]{akata2020research}
Zeynep Akata, Dan Balliet, Maarten de~Rijke, Frank Dignum, Virginia Dignum,
  Guszti Eiben, Antske Fokkens, Davide Grossi, Koen Hindriks, Holger Hoos,
  et~al.
\newblock A research agenda for hybrid intelligence: Augmenting human intellect
  with collaborative, adaptive, responsible, and explainable artificial
  intelligence.
\newblock \emph{Computer}, 53\penalty0 (8):\penalty0 18--28, 2020.

\bibitem[Alegre et~al.(2016)Alegre, Augusto, and Clark]{alegre2016engineering}
Unai Alegre, Juan~Carlos Augusto, and Tony Clark.
\newblock Engineering context-aware systems and applications: A survey.
\newblock \emph{Journal of Systems and Software}, 117:\penalty0 55--83, 2016.

\bibitem[Baumgartner and Retschitzegger(2006)]{baumgartner2006survey}
Norbert Baumgartner and Werner Retschitzegger.
\newblock A survey of upper ontologies for situation awareness.
\newblock In \emph{Proc. of the 4th IASTED International Conference on
  Knowledge Sharing and Collaborative Engineering}, pages 1--9, 2006.

\bibitem[Doshi-Velez and Kim(2017)]{doshi2017towards}
Finale Doshi-Velez and Been Kim.
\newblock Towards a rigorous science of interpretable machine learning.
\newblock \emph{arXiv preprint arXiv:1702.08608}, 2017.

\bibitem[Endsley(1995)]{endsley1995toward}
Mica~R Endsley.
\newblock Toward a theory of situation awareness in dynamic systems.
\newblock \emph{Human factors}, 37\penalty0 (1):\penalty0 32--64, 1995.

\bibitem[Kayal et~al.(2018)Kayal, Brinkman, Neerincx, and
  Riemsdijk]{kayal2018automatic}
Alex Kayal, Willem-Paul Brinkman, Mark~A Neerincx, and M~Birna~Van Riemsdijk.
\newblock Automatic resolution of normative conflicts in supportive technology
  based on user values.
\newblock \emph{ACM Transactions on Internet Technology}, 18\penalty0
  (4):\penalty0 1--21, 2018.

\bibitem[Kola et~al.(2019)Kola, Jonker, and van Riemsdijk]{kola2019s}
Ilir Kola, Catholijn~M Jonker, and M~Birna van Riemsdijk.
\newblock Who's that? social situation awareness for behaviour support agents.
\newblock In \emph{International Workshop on Engineering Multi-Agent Systems},
  pages 127--151. Springer, 2019.

\bibitem[Kola et~al.(2020{\natexlab{a}})Kola, Jonker, Tielman, and van
  Riemsdijk]{kola2020grouping}
Ilir Kola, Catholijn~M Jonker, Myrthe~L Tielman, and M~Birna van Riemsdijk.
\newblock Grouping situations based on their psychological characteristics
  gives insight into personal values.
\newblock In \emph{11th International Workshop Modelling and Reasoning in
  Context}, pages 17--26, 2020{\natexlab{a}}.

\bibitem[Kola et~al.(2020{\natexlab{b}})Kola, Tielman, Jonker, and van
  Riemsdijk]{kola2020predicting}
Ilir Kola, Myrthe~L Tielman, Catholijn~M Jonker, and M~Birna van Riemsdijk.
\newblock Predicting the priority of social situations for personal assistant
  agents.
\newblock In \emph{PRIMA}, 2020{\natexlab{b}}.

\bibitem[Kola et~al.(2021)Kola, Jonker, and van Riemsdijk]{kola2021predicting}
Ilir Kola, Catholijn~M Jonker, and M.~Birna van Riemsdijk.
\newblock Using psychological characteristics of situations for social
  situation comprehension in support agents.
\newblock \emph{arXiv preprint arXiv:2110.09397}, 2021.

\bibitem[Lewin(1939)]{lewin1939field}
Kurt Lewin.
\newblock Field theory and experiment in social psychology: Concepts and
  methods.
\newblock \emph{American journal of sociology}, 44\penalty0 (6):\penalty0
  868--896, 1939.

\bibitem[Murukannaiah and Singh(2015)]{murukannaiah2015platys}
Pradeep~K Murukannaiah and Munindar~P Singh.
\newblock Platys: An active learning framework for place-aware application
  development and its evaluation.
\newblock \emph{ACM Transactions on Software Engineering and Methodology},
  24\penalty0 (3):\penalty0 1--32, 2015.

\bibitem[Rauthmann et~al.(2014)Rauthmann, Gallardo-Pujol, Guillaume, Todd,
  Nave, Sherman, Ziegler, Jones, and Funder]{rauthmann2014situational}
John~F Rauthmann, David Gallardo-Pujol, Esther~M Guillaume, Elysia Todd,
  Christopher~S Nave, Ryne~A Sherman, Matthias Ziegler, Ashley~Bell Jones, and
  David~C Funder.
\newblock The situational eight diamonds: A taxonomy of major dimensions of
  situation characteristics.
\newblock \emph{Journal of Personality and Social Psychology}, 107\penalty0
  (4):\penalty0 677, 2014.

\bibitem[Rosatelli et~al.(2019)Rosatelli, Gedik, and
  Hung]{rosatelli2019detecting}
Alessio Rosatelli, Ekin Gedik, and Hayley Hung.
\newblock Detecting f-formations \& roles in crowded social scenes with
  wearables: Combining proxemics \& dynamics using lstms.
\newblock In \emph{2019 8th International Conference on Affective Computing and
  Intelligent Interaction Workshops and Demos (ACIIW)}, pages 147--153. IEEE,
  2019.

\bibitem[Schwartz(1992)]{schwartz1992universals}
Shalom~H Schwartz.
\newblock Universals in the content and structure of values: Theoretical
  advances and empirical tests in 20 countries.
\newblock \emph{Advances in experimental social psychology}, 25\penalty0
  (1):\penalty0 1--65, 1992.

\bibitem[Yang et~al.(2009)Yang, Read, and Miller]{yang2009concept}
Yu~Yang, Stephen~J Read, and Lynn~C Miller.
\newblock The concept of situations.
\newblock \emph{Social and Personality Psychology Compass}, 3\penalty0
  (6):\penalty0 1018--1037, 2009.

\end{thebibliography}

\end{document}